\documentclass[letterpaper]{article} 
\usepackage{aaai24}  
\usepackage{times}  
\usepackage{helvet}  
\usepackage{courier}  
\usepackage[hyphens]{url}  
\usepackage{graphicx} 

\usepackage{natbib}  
\usepackage{caption} 
\nocopyright
\usepackage{hyperref}
\usepackage{booktabs}
\usepackage{amsmath}
\usepackage{amssymb}
\usepackage{pifont}

\newcommand{\plm}{P_{\text{LM}}}
\newcommand\T{\rule{0pt}{2.2ex}}       
\newcommand\B{\rule[-1.0ex]{0pt}{0pt}} 
\newcommand{\cmark}{\ding{51}}
\newcommand{\xmark}{\ding{55}}

\def\secref#1{Section~\ref{#1}}
\def\figref#1{Fig.~\ref{#1}}
\def\tabref#1{Table~\ref{#1}}
\def\eqref#1{(\ref{#1})}

\title{Discrete Prompt Compression with Reinforcement Learning}

\author{
    Hoyoun Jung\textsuperscript{\rm 1},
    Kyung-Joong Kim, \textsuperscript{\rm 1}
}
\affiliations{
    \textsuperscript{\rm 1}School of Integrated Technology\\
    Gwangju Institute of Science and Technology\\
    123 Chem-dan gwa-gi-ro, Gwangju 61005, Korea\\
    nenomigami@gm.gist.ac.kr, kjkim@gist.ac.kr
}

\begin{document}

\maketitle

\begin{abstract}
 Compressed prompts aid instruction-tuned language models (LMs) in overcoming context window limitations and reducing computational costs. Existing methods, which are primarily based on training embeddings, face various challenges associated with interpretability, the fixed number of embedding tokens, reusability across different LMs, and inapplicability when interacting with black-box APIs. This study proposes prompt compression with reinforcement learning (PCRL), which is a discrete prompt compression method that addresses these issues. The proposed PCRL method utilizes a computationally efficient policy network that edits prompts directly. The training approach employed in the proposed PCRLs can be applied flexibly to various types of LMs, including both decoder-only and encoder-decoder architecture and it can be trained without gradient access to the LMs or labeled data. The proposed PCRL achieves an average reduction of 24.6\% in terms of the token count across various instruction prompts while maintaining sufficient performance. In addition, we demonstrate that the learned policy can be transferred to larger LMs, and through a comprehensive analysis, we explore the token importance within the prompts. Our code is accessible at \href{https://github.com/nenomigami/PromptCompressor}{https://github.com/nenomigami/PromptCompressor}.
\end{abstract}

\section{Introduction}
Instruction-tuned language models (LMs) \cite{wei2021finetuned,ouyang2022training,sanh2022multitask}, e.g., ChatGPT, are being used increasingly to address various natural language processing (NLP) challenges, offering solutions through task-specific prompts for both individuals and businesses. The design of concise prompts that contain only essential information benefits both users and servers. For example, Users benefit from reduced query-length dependent API usage costs and overcoming context window limitations, and servers benefit from shorter prompt designs that reduce computational burden. Prompt compression methods for concise, information-rich prompts are beneficial in terms of realizing efficient LM utilization.

A widely adopted prompt compression method involves training embeddings that encapsulate the original contexts \cite{wingate2022prompt,mu2023learning,chevalier2023adapting}, using the \textit{soft} prompt concept \cite{lester2021power}. However, with this method, the appropriate embedding token count must be determined, its inherent properties can hinder interpretation, it lacks cross-model reusability, and its dependency on gradient access to LMs can make it impractical for scenarios that employ API services. An appealing alternative is compression via \textit{discrete} prompts that comprise concrete tokens from the vocabulary. Only a few studies have investigated methods to compress discrete prompts. One such study, the selective-context by \cite{li2023unlocking}, focuses on reducing prompt length by filtering out less informative text based on self-information from an entropy perspective.

\begin{table*}[t]
\centering
\begin{tabular}{cccccccc}
\toprule
\textbf{Method}            & \textbf{Generalization} & \textbf{\begin{tabular}[c]{@{}l@{}} \quad Adpative \\ Compression\end{tabular}} \ & \; \textbf{\begin{tabular}[c]{@{}l@{}}Black-Box \; \\  Applicable\end{tabular}} \; & \textbf{\begin{tabular}[c]{@{}l@{}}Transferrable\\ \quad b/w LMs\end{tabular}} &  \textbf{Interpretability} \\ \midrule
Fine Tuning                & \xmark   & \xmark   & \xmark   & \xmark   & \xmark           \\
Wingate et al., (2022)     & \xmark   & \xmark   & \xmark   & \xmark   & \xmark           \\
Mu et al., (2023)          & \cmark   & \xmark   & \xmark   & \xmark   & \xmark           \\
Chevalier et al., (2023)   & \cmark   & \cmark   & \xmark   & \xmark   & \xmark           \\ 
\midrule
\textbf{PCRL (Ours)}      & \cmark   & \cmark   & \cmark   & \cmark   & \cmark           \\ \bottomrule
\end{tabular}
\caption{
Comparison of the proposed model with soft prompt compression methods based on selected desirable properties. Generalization represents the characteristic that allows it to handle new prompts without requiring retraining. A model that is capable of adaptive compression adjusts the length of the compressed prompt according to the length of the original prompt. Black-box applicable methods can be applied in black-box API scenarios where gradient or token probability are not provided. Our model demonstrates transferability between various LMs by using discrete tokens rather than embeddings (\secref{analysis}).}
\label{proscons}
\end{table*}

In this paper, we proposed the prompt compression with reinforcement learning (PCRL) method that utilizes a discrete prompt compression technique that incorporates the advantages outlined in \tabref{proscons}. Drawing on techniques similar to those used for extractive summarization tasks, the learned policy edits prompts directly, which reduces tokens with limited contribution to the LM output (i.e., the generation LM). To reduce the computational overhead associated with the compression process, we designed a process that determines the inclusion or exclusion of each token simultaneously in a single step. In addition, the policy integrates MLP layers with a small number of parameters into lightweight LMs (i.e., the policy LM), which improves computational efficiency further.

The model is trained by a reward function that balances both the faithfulness of the compressed prompts and their reduced length using a policy gradient algorithm \cite{sutton1999policy}. Here, faithfulness is evaluated indirectly by measuring the similarity between the output of the generation LMs when given uncompressed and compressed prompts. This approach allows us to train the policy without the gradients of the LMs and ensures effective learning even in the absence of data labels. In addition, this enables consistent training regardless of whether the generation LM has a decoder-only or encoder-decoder architecture.

The proposed model achieved an average compression ratio of 24.6\% in experiments conducted on various instruction sets while maintaining output quality that is similar to that of the original prompts. In addition, we analyzed the importance of tokens for the response and the results provide insights that could be used to further refine and optimize the compression technique. Furthermore, we found that the policy learned from a smaller model can potentially be transferred to larger and more powerful generation LMs.

The primary contributions of this study are summarized as follows:
\begin{itemize}
    \item We propose the discrete prompt compression concept and describe the problem using RL.
    \item We demonstrate the superior performance of the proposed PCRL method compared to existing methods and the transferability of the learned policy to more practical LMs.
    \item We explore the token characteristics within the prompt that yield minimal contribution to the LM output.
\end{itemize}

\section{Related Work}
\subsection{Discrete Prompt Optimization}
Prompting has been widely used as a general method for NLP tasks \cite{brown2020language,schick2021exploiting,sanh2022multitask}, and corresponding research into prompt optimization in LMs has emerged as a significant area of study. For example, prompt tuning optimizes continuous embeddings using gradient descent \cite{lester2021power,liu2021gpt}. In contrast, discrete prompt optimization searches for tokens or exemplars to construct effective prompts. A previous study by \citet{shin2020autoprompt} utilized gradient information to search for the best performing prompt, and another study \cite{prasad2023grips} proposed an edit-based search method that is applicable to gradient-free scenarios. In addition, \citet{zhou2022large} leveraged LMs to generate and evaluate prompts. \citet{deng2022rlprompt} introduced an RL-based framework to generate optimal prompts and improve LM performance. \citet{zhang2022tempera} integrated various prompt components, including exemplars and the verbalizer, which were optimized using RL. These studies have made remarkable progress; however, they focused on enhancing performance, largely neglecting the prompt compression perspective.

\subsection{Prompt Compression}
In the prompt compression research field, the majority of the studies adopt a soft prompts concept. Early studies set distillation objectives to minimize the discrepancies between the generative distributions produced by LMs using original prompts and those produced using soft prompts \cite{wingate2022prompt}. However, this technique requires re-optimization for each new prompt, thereby lacking the capability to generate compressed prompts for different prompts. \citet{mu2023learning} decomposed a prompt into tasks and inputs, which effectively reduced the task component to a few gist tokens. The proposed method differs in that it attempts to compress the entire prompt. \citet{chevalier2023adapting} focused on overcoming limited context windows using compressed summary vectors from long contexts. Similar to our work, \citet{li2023unlocking} removed less informative content from discrete prompts by calculating self-information based on the likelihood of tokens. 
However, this method is dependent on having access to probability information, which is unfeasible in black-box API scenarios. 

\subsection{Unsupervised Summarization}
A different perspective of the proposed study involves unsupervised summarization to create more concise prompts. Specifically, we select an extractive summarization over abstractive methods to reduce the search space and maintain closer context with the original prompt. \citet{zhou2019simple} employed a pretrained model with the beam search technique to identify tokens that maximize both fluency and similarity. \citet{schumann2020discrete} used a greedy hill-climbing search strategy to optimize objectives for fluency and similarity. In addition, \citet{narayan2018ranking} implemented extractive summarization through RL using ROUGE \cite{lin2003automatic} scores in the design of the reward function. Similarly, \citet{ghalandari2022efficient} trained an extractive policy that receives rewards based on fluency, similarity, and length metrics. Most of these studies summarized content based on the source text; however, the proposed method is distinguished by its use of responses generated from the LMs through prompts.

\section{Prompt Compression with RL}
\subsection{Task}
Here, given a prompt $p=\{x_1, x_2, ..., x_n\}$, comprising tokens $x_i$, a compressed prompt $p'$ is defined as a shorter sequence of tokens. When input to LMs, it produces a generative distribution $\plm(\cdot|p')$ that is similar to that obtained by the original prompt $\plm(\cdot|p)$. The output sequence of tokens is denoted $y$, and the function $\delta$ quantifies the divergence between the distributions. The compressed prompt should satisfy the following condition.

\begin{align}
\delta(\plm(y|p), \plm(y|p')) < \epsilon, \;\   |p'| < |p|
\end{align} 

The primary objective of this study is to learn a policy $\pi$ that compresses a given original prompt $p$ as much as possible. When applied to a prompt $p$, this policy generates a shorter prompt $p^\pi = \pi(p)$ that retains the semantic information of $p$. We cast this problem as a sequence labeling task to select salient tokens from the prompt. In this context, an include/exclude label is assigned for each token $x_i$, thereby creating a compressed prompt that encompasses only the required tokens. The optimization objective of this policy combines two terms, i.e., faithfulness and the compression ratio, using the balance term $\beta$.

\begin{align}
\label{obj}
\pi^* = \arg\min_{\pi}[ \ \delta(\plm(y|p), \plm(y|p^\pi)) + \beta\left(|p^\pi| / |p|\right) \ ]
\end{align}

Typically, common methods that use the soft prompt fix the token length of the compressed prompt as a hyperparameter and minimize the divergence $\delta$ as a loss through gradient descent. However, challenges arise when practitioners interact with LMs via an API or when computing the gradients becomes excessively costly. This frequently makes it unfeasible to access the probability distribution of output tokens $\plm(\cdot|p)$ and the gradient information directly. To overcome this specific challenge, we reformulate the problem using RL, by leveraging optimization without the LM gradient. In addition, we replace the measure of divergence between the output distributions $\plm(\cdot|p)$ with a measure of similarity of the output sequences $y = LM(p)$. In addition, we adopt the ROUGE score to compute similarity in the proposed model. 

\subsection{Training Procedure}

The construction of the compressed prompts is formulated as a discrete prompt optimization problem, which we address using RL. To accomplish this, we set up the following Markov decision process (MDP). Given an initial state, i.e., tokenized prompt $p = \{x_1, x_2,\dots,x_n\}$, the policy $\pi$ outputs binary labels as actions $a = \{a_1, a_2,\dots,a_n\} \in \{0, 1\}^n$ for each token. Here, each label $a_i$ determines whether the corresponding token is included or excluded. Although this method may yield grammatically inconsistent prompts, a recent study suggests they could be more effective \cite{deng2022rlprompt}. Following the transition to a compressed prompt $p^\pi$, a reward $R(p,a)$ is received. This reward is calculated from the output sequences of the LMs and the reduced prompt length. Note that the MDP terminates in a single step, thus, our environment resembles a contextual multi-armed bandit \cite{lu2010contextual}. In contrast to the traditional bandit problem, in which only a single action and its corresponding reward are available in each episode, our algorithm allows the policy to obtain rewards with multiple possible actions.

\begin{figure}[t]
\centering
\includegraphics[width=0.9\columnwidth]{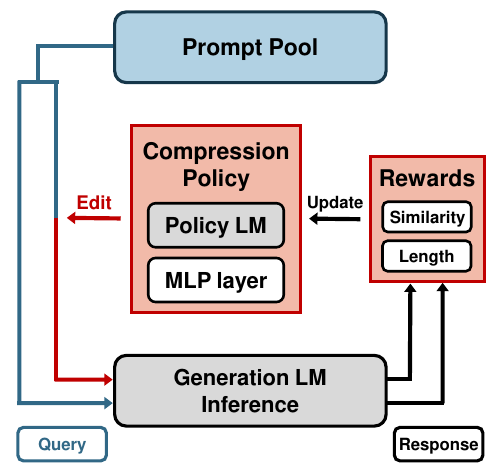}
\caption{Overall training procedure of PCRL. A prompt is sampled from the prompt pool, edited by the compression policy, and evaluated by comparing the generation LM's response to the original and edited prompt. The resulting reward is used for policy updates.}
\label{procedure}
\end{figure}
\figref{procedure} illustrated the training procedure of the proposed method. First, a prompt $p$ is sampled randomly from the prompt pool, which is a dataset of prompts that do not require labels. The sampled prompt is processed through the compression policy $\pi_\theta$ to produce a compressed prompt $p^\pi$. The original and compressed prompts are then input to the LMs, yielding two sets of output responses. Then the reward is calculated based on the measured similarity and the compression ratio of $p^\pi$. To balance accuracy and time efficiency during the generation process, we limit the number of generated tokens to $T$. Note that a longer and more time-consuming generation process could offer more accurate understanding of the similarity, However, empirical findings indicate that even a partial generation is sufficient. 

The compression policy $\pi_\theta$ (parameterized by $\theta$) is trained using the policy gradient algorithm. This process ensures that, given an input prompt $p$, the policy will yield a probability distribution of binary actions $a_i$ for each token. Here, the objective is to identify the parameter $\theta$ that causes $\pi_\theta(a|p)$ to assign a high preservation probability to tokens that convey the essence of the prompts, which is accomplished by maximizing the following objective function in relation to the parameters $\theta$.

\begin{align}
J(\theta) = E_{a\sim\pi_\theta}[R(p, a)]
\end{align} 
\noindent where, $\pi_\theta$ stands for $\pi_\theta(a|p)$. The policy gradient algorithm possesses the following gradient:

\begin{align}
\nabla_\theta J(\theta) = E\left[R(p, a) \ \nabla_\theta \  \text{log} \ \pi_\theta(a|p)\right]
\end{align} 

Note that we subtract a baseline from the reward to facilitate effective learning by adopting Self-critical sequence training (SCST) \cite{rennie2017self}. For this training algorithm, $R(p, a)$ is the general reward obtained by executing the action $a$ sampled from the current policy $\pi(\cdot|p)$. Conversely, the baseline $R(p, \hat{a})$ is derived by executing the action $\hat{a}$ with the highest probability in the current policy.
\begin{align}
\nabla_\theta J(\theta) = E\left[(R(p, a) - R(p, \hat{a}))\nabla_\theta \ \text{log} \ \pi_\theta(a|p)\right]
\end{align} 
In simpler terms, this implies that an action is considered preferable if it offers a reward that is greater than that predicted by the current policy. Both the ROUGE scores and compression ratio, which are used as reward functions, are positive; thus, it is necessary to penalize actions that yield relatively lower rewards. Incorporating the baseline helps us deal with this concern effectively.

\begin{figure}[t]
\centering
\includegraphics[width=0.90\columnwidth]{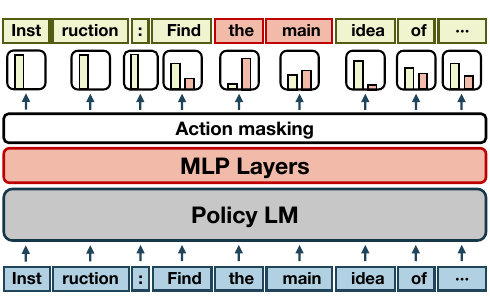}
\caption{The policy network of PCRL. When a tokenized prompt is inputted, the network outputs an include/exclude probability for each token. If a token is the part of a statement, the exclude action is masked out.}
\label{model}
\end{figure}

A limitation of SCST occurs when the two sequences achieve comparable rewards $R(p, a) \simeq R(p, \hat{a}) $, i.e., the loss approaches zero, and the model has little to learn, thereby wasting a training sample \cite{laban2021keep}. Thus, we enhance the learning process by sampling $k$ actions from the current policy for the same prompts and calculating the average rewards. In addition, to reduce instances where the loss is near zero, we implement an entropy term to the loss, which increases the probability of sampling diverse actions \cite{haarnoja2017reinforcement}. We then train the model by minimizing the following loss function. 
\begin{align}
\mathcal{L}(\theta) =  (R(p, a) - R(p, \hat{a})) \ \text{log} \ \pi_\theta(a|p) + 
\alpha \mathcal{H}(\pi(\cdot|p))
\end{align} 
Here, the temperature parameter $\alpha$ determines the significance of the entropy term.


\subsection{Model Architecture}

\figref{model} shows the architecture of the policy network $\pi_\theta$. Here, We attach binary classification MLP layers to a frozen pre-trained policy LM, which is used to extract the contextual embeddings of the tokens. A primary motivation behind compressing prompts is the need to reduce computational costs, leading us to favor efficient, smaller-scale backbone models, e.g., DistilBERT \cite{sanh2019distilbert}. During training, only the parameters in the attached simple MLP layers are updated. We use action masks to prevent the policy from excluding statement tokens (e.g., ``Instruction: " and ``Input: ") to ensure that the compression ratio reflects the actual reduction of the prompt rather than simply removing these statement tokens. In addition, the policy LM does not necessarily have to be the same as the generation LM for which we optimize the prompt. 

\subsection{Reward Design}
Note that the reward function must balance two potentially conflicting terms, i.e., faithfulness and reduced length. To account for faithfulness, we define a term based on the ROUGE-L score of two output token sequences generated from the original prompt $p$ and the compressed prompt $p^\pi$.

\begin{align}
R_f = \text{ROUGE-L}(LM(p), LM(p^\pi))
\end{align}

\noindent The ROUGE-L score considers sentence-level structural similarity; thus it is suitable as a faithfulness term. To reflect the reduced length, we use the compression ratio, which is the proportion of the reduced token count to the original token count in the prompts. The final reward is given as follows.

\begin{align}
R(p, a) &=
\begin{cases}
1 - |p^\pi| / |p| & \text{if} \; R_f \geq \tau \\
-\lambda & \text{else}
\end{cases}
\end{align}
\noindent If the ROUGE-L score exceeds a certain threshold $\tau$, the model receives the compression ratio as the reward; however, if the score does not exceed threshold $\tau$, the model receives a penalty $\lambda$.

A key difference between the proposed method and typical RL-based summarization \cite{laban2020summary,ghalandari2022efficient} is that we do not consider grammatical correctness. Recent studies \cite{webson2022prompt,prasad2023grips,deng2022rlprompt} have suggested that LMs leveraging prompts do not necessarily adhere to human language patterns. 
Interestingly, prompts that yield high performance tend to be gibberish without a clear human-understandable meaning. Thus, we do not incorporate grammatical fluency into the reward function. In fact, this aspect facilitates the potential to acquire shorter prompts.

\begin{table*}[t]
\centering
\small{
\begin{tabular}{cccccccccc}
\toprule
                    & \multicolumn{3}{c}{\textbf{Seen}}       & \multicolumn{3}{c}{\textbf{Unseen}}     & \multicolumn{3}{c}{\textbf{Human}}      \\
                    & ROUGE-L    & ChatGPT \% & Cr   & ROUGE-L    & ChatGPT \% & Cr   & ROUGE-L    & ChatGPT \% & Cr   \\ \toprule
GPT2-XL             &            &            &      &            &            &      &            &            &      \\ \midrule 
Original            & 54.5 (100) & 50.0 (100) & 0.0  & 44.5 (100) & 50.0 (100) & 0.0  & 23.2 (100) & 50.0 (100) & 0.0  \\
w/o Stopwords       & 38.0 (69.6)& 36.6 (73.2)& 33.9 & 35.1 (79.0)& 38.6 (77.2)& 30.9 & 18.0 (77.6)& 40.2 (80.4)& 34.5 \\
Selective Context   & 46.4 (85.0)& 41.1 (82.2)& 21.9 & 40.9 (91.9)& 42.5 (85.0)& 22.4 & 20.1 (86.9)& 45.8 (91.6)& 21.6 \\
\textbf{PCRL (Ours)}& \textbf{51.0 (93.6)}& \textbf{47.3 (94.6)}& 21.8 &\textbf{ 42.3 (95.1)}& \textbf{49.1 (98.2)}& 23.2 & \textbf{20.5 (88.6)}& \textbf{47.1 (94.3)}& 24.3
\\ \midrule
FLAN-T5-XL          &            &            &      &            &            &      &            &            &      \\ \midrule
Original            & 44.3 (100) & 50.0 (100) & 0.0  & 43.7 (100) & 50.0 (100) & 0.0  & 23.3 (100) & 50.0 (100) & 0.0  \\
w/o Stopwords       & 34.8 (78.5)& 40.7 (81.4)& 32.6 & 36.3 (83.0)& 40.0 (80.0)& 29.7 & 19.4 (83.2)& 39.7 (79.4)& 32.8 \\
Selective Context   & 37.3 (84.2)& 38.6 (77.2)& 24.9 & 38.1 (87.2)& 38.8 (77.6)& 25.1 & 19.8 (84.5)& 36.1 (72.2)& 25.4 \\
\textbf{PCRL (Ours)}& \textbf{41.1 (92.9)}& \textbf{45.0 (90.0)}& 27.4 & \textbf{40.6 (92.8)}& \textbf{43.6 (87.2)}& 25.1 & \textbf{21.1 (90.5)}& \textbf{41.9 (83.8)}& 27.6 \\ 
\bottomrule
\end{tabular}
}
\caption{
ROUGE-L and ChatGPT performance of PCRL for instruction prompts. Values in parentheses indicate normalized scores to the Original.
}
\label{comptable}
\end{table*}

\section{Experiments}
Through a series of experiments, we demonstrate that the proposed PCRL method compresses prompts successfully regardless of the type of the generation LMs. In these experiment, we fine-tuned the LMs using a diverse set of instruction data to mimic off-the-shelf instruction-tuned LMs. We then evaluated the performance of the compressed prompts obtained by the PCRL method on a validation instruction set. In addition, the experimental results demonstrate that the transferability of the compression policy across LMs allows us to learn from smaller models in a cost-effective manner and apply it to larger, more powerful models.

\subsection{Instruction Prompts}
\label{instruction}
\subsubsection{Datasets}
To construct LMs that can be generalized across various instructions, we used the Alpaca+ dataset, following a previous study \cite{mu2023learning}. The Alpaca+ dataset consists of a Self-instruct \cite{wang2022self} and a Stanford Alpaca \cite{alpaca} dataset. Specifically, it comprises (\textit{tasks}, \textit{input}, \textit{answer}) tuples, with a total of 104,664 unique tasks, and it is effective for experiments involving a diverse set of instructions. The validation set in the Alpaca+ dataset is categorized into three distinct sets. The first set, Seen prompts, contains 1,000 prompts in which the tasks are already seen in the training set; however, the inputs are new.
The second set, Unseen prompts, includes 1,000 prompts where both the tasks and the inputs have never been encountered in the training set. The final set includes 252 handcrafted human prompts, thereby representing a substantial out-of-distribution (OOD) challenge.

\subsubsection{Models}
In these experiments, we employed two different architectures to demonstrate that the proposed method can be applied to various text generation LMs. The first LM is GPT2-XL \cite{radford2019language}, which is a decoder-only model, and the second LM is FLAN-T5-XL \cite{chung2022scaling}, which is an encoder-decoder model. These LMs include 1.5B and 3.0B parameters, respectively. Each of these models was fine-tuned on the Alpaca+ dataset with three epochs for GPT2-XL and one epoch for FLAN-T5-XL to create instruction-tuned models for inference. The performance achieved with noncompressed prompts, which are used as the upper-bound baseline original, is the standard for evaluating our models. 

Several approaches were considered for comparison. including the basic technique of eliminating less informative tokens (specifically stop words) using the NLTK stop word list \cite{bird2009natural}.  In addition, we compared the proposed model's effectiveness with that of the selective-context method \cite{li2023unlocking}. To ensure fairness in the comparison, we configured the model to perform compression at the token level with similar compression ratios and maintained the inclusion of statement tokens. We evaluated both the foundation model and the instruction-tuned model to calculate self-information, and we reported the best obtained results.

\subsubsection{Evaluation}
The evaluation metrics used to assess model performance included ROUGE-L, ChatGPT, and compression metrics. ROUGE-L has been used in instruction fine-tuning \cite{wei2021finetuned,wang2022super} and prompt compression \cite{mu2023learning} studies. 
ROUGE-L calculates the similarity between the ground truth (Gt) and the generated response (Gen) by measuring F1 score of the longest common subsequence (LCS). This similarity is quantified using the following formulas:
\begin{align}
P_{LCS} &= {\text{LCS}(\text{Gen}, \text{Gt})}\ /\ {|\text{Gen}|} \\
R_{LCS} &= {\text{LCS}(\text{Gen}, \text{Gt})}\ /\ {|\text{Gt}|}
\end{align}
\vspace{-5mm} 
\begin{align}
\text{ROUGE-L}(\text{Gen}, \text{Gt}) = \frac{2 \times P_{LCS} \times R_{LCS}}{P_{LCS} + R_{LCS}}
\end{align}
\noindent It is important to distinguish this usage from that in the reward calculation. The reward function employs ROUGE-L to calculate the score by comparing the sentences generated from the original and compressed prompts; however, during evaluation, it represents the similarity to the true reference in the dataset. GPT2-XL tends to continue generating tokens until it reaches the maximum token limit; thus, we generate tokens up to the number of tokens in the reference sentences for both models.

The compression ratio (Cr) is the reduced token count in the compressed prompt divided by the token count in the original prompt. 
\begin{align}
\text{Cr} &= 1 - {|p^\pi|}\ / \ {|p|}
\end{align}
To ensure fairness, we calculate Cr by excluding the number of statement tokens. This ratio signifies the model's effectiveness in terms of condensing the original prompt. Due to potential differences between the tokenizers used by the policy and the generation LMs, we employ the decoded text as a bridge. Here, tokens are edited on the basis of the policy LM's tokenizer, and the Cr is calculated using the generation LM's tokenizer.

The ChatGPT metric represents the ratio by which ChatGPT selects the better response between two options for a given task. Here, the objects of comparison are the responses to our model's compressed prompt and the original prompt.
\begin{align}
ChatGPT = \frac{\#\, \text{LM}(p^\pi) \text{ selected as better than LM}(p)}{\#\ \text{Comparisons}}
\end{align}
The ChatGPT metric can be used as a supplement because it can consider more semantic elements than the ROUGE-L metric. If the compressed prompt is similar in meaning to the original prompt, a result approximating 50\% is expected. This metric is considerably faster and more cost-effective than human evaluation; however, it exhibits nearly the same performance as human annotators in instruction-following tasks \cite{mu2023learning}. In addition, the near-human performance of ChatGPT in text annotation and evaluation \cite{gilardi2023chatgpt,huang2023chatgpt,wang2023chatgpt}, lends credibility to this measure. The prompt given to ChatGPT follows precisely that described in the literature \cite{mu2023learning} without any additional prompt engineering.

\subsubsection{Results}

The experimental results for the instruction-following tasks on the entire validation set are shown in Table 2. As can be seen, the proposed model outperformed the compared methods on all validation sets. For the GPT2-XL model, our compression policy achieved performance similar to that of the original prompts' ROUGE-L scores and the ChatGPT metrics across most validation sets. This was achieved while also reducing the number of input tokens by an average of 22.7\% for GPT2-XL and 26.4\% for FLAN-T5-XL. In the human split set, both the ROUGE-L scores and the ChatGPT metrics exhibited lower overall values. In this split, it appears that the OOD challenge makes it difficult for the policy to compress considering the context.


\begin{table}
\begin{tabular}{ccccc}
\toprule
Generation                                                        & \multicolumn{2}{c}{GPT2-XL} & \multicolumn{2}{c}{FLAN-T5-XL}               \\
Model (Size)                                                      & ChatGPT \%                  & Cr       & ChatGPT \%         & Cr           \\ \midrule    
\begin{tabular}[c]{@{}c@{}} Falcon-\\ instruct (7B)\end{tabular}  & 43.7 \small{($\pm2.0$)}     & 22.0     & 42.2 \small{($\pm2.1$)} & 26.9    \\ 
\begin{tabular}[c]{@{}c@{}} LLaMa2-\\ chat-hf (7B)\end{tabular}    & 47.3 \small{($\pm1.8$)}     & 21.7     & 45.8 \small{($\pm1.9$)} & 26.5    \\ 
\begin{tabular}[c]{@{}c@{}} FLAN-T5-\\ XXL (11B)\end{tabular}      & 44.8 \small{($\pm2.3$)}     & 22.2     & 42.7 \small{($\pm1.5$)} & 26.8    \\ 
\begin{tabular}[c]{@{}c@{}} GPT-3.5\\-Turbo\end{tabular}           & 49.8 \small{($\pm1.1$)}     & 21.8     & 47.7 \small{($\pm1.1$)} & 27.0    \\ \bottomrule
\end{tabular}
\caption{
Transferability of the proposed PCRL method across different LMs, evaluated using the ChatGPT metric. Values in parentheses indicate the 95\% confidence interval.
}
\label{transf}
\end{table}

\subsection{Transferring Prompts across LMs}
A unique advantage of discrete prompts over soft prompts is that they are transferrable across models because of the common text space rather than the model-specific latent space \cite{deng2022rlprompt}. Leveraging this advantage, we demonstrate the practicality of the proposed model by experimenting with its application to larger, more powerful generation LMs. The results of this experiment effectively prove that the proposed method's use of discrete prompts enables higher flexibility and robustness, thereby making it a valuable tool in various scenarios and across different models.

\subsubsection{Experiment}
We evaluated the transfer ability of the proposed method using 2,252 data points, which is the sum of all validation sets used in the previous experiment. Here, we considered four models, i.e., LLaMa2 \cite{touvron2023llama}, which is a decoder-only model with 7B parameters, Falcon \cite{almazrouei2023falcon}, which is another decoder-only model with 7B parameters, FLAN-T5-XXL \cite{chung2022scaling}, which is an encoder-decoder architecture with 11B parameters and GPT-3.5 model, which is the LM used in ChatGPT. Specifically, we used the \texttt{Llama-2-7B-chat-hf}, \texttt{Falcon-7B-instruct}, \texttt{FLAN-T5-XXL} and \texttt{gpt-3.5-turbo} models without fine-tuning. In line with our previous experiments, we compared the output of the original and compressed prompts using the ChatGPT metric. This allowed us to effectively assess how well the proposed method performs across different models, by showcasing its flexibility and potential for adaptation to various scenarios.

\subsubsection{Results}
\tabref{transf} shows the transfer results for compression policies applied to various large LMs. These policies were trained using GPT2-XL and FLAN-T5-XL as the generation LMs. As can be seen, the difference in the compression ratio due to variations in the tokenizers between the generation LMs was minimal. As a result, the Cr value was similar to that obtained in the previous experiments.

Surprisingly, we found that the ChatGPT evaluation is generally consistent with the results of the original generation models, and in some cases, it even surpasses them. Specifically, LLaMa2 demonstrated a successful transfer with a win rate of 47.3\% in the GPT2-XL model and 45.8\% in the FLAN-T5-XL model. In addition, the performance of GPT-3.5 surpassed the result obtained by models used in training, achieving 49.8\% in the GPT2-XL model and 47.7\% in the FLAN-T5-XL model. 

The level of stability emphasizes the viability of the proposed method, indicating its effectiveness even with updates to the API version or when an entirely different LM is used.

The results from LLaMa2 and GPT-3.5 suggest the possibility that the more powerful the model, the less susceptible it is to the influence of redundant tokens, thereby indicating a higher potential for compression. In addition, the performance of the FLAN-T5-XXL model lagged behind the other models, despite employing the same training procedure and tokenizer as the FLAN-T5-XL model. This variation may stem from the fine-tuning differences on the Alpaca+ dataset, causing a deviation from the performance observed with the original FLAN-T5-XL model.

\begin{table}[tbp]
\centering
\begin{tabular}{|ccc|ccc|}
\toprule
Token & \begin{tabular}[c]{@{}c@{}}Freq \\ Rank\end{tabular} & \begin{tabular}[c]{@{}c@{}}Removal \\ Ratio\end{tabular} & Token & \begin{tabular}[c]{@{}c@{}}Freq \\ Rank\end{tabular} & \begin{tabular}[c]{@{}c@{}}Removal\\ Ratio\end{tabular} \\ \midrule
ribe  & 102          & 99.97           & mine  & 732           & 91.54                             \\
ify   & 61           & 99.96           & ated  & 535           & 90.60                             \\
.     & 3            & 98.70           & me    & 40            & 89.80                             \\
a     & 10           & 97.86           & of    & 11            & 89.23                             \\
ize   & 266          & 97.40           & them  & 138           & 85.81                             \\
.,    & 588          & 95.65           & an    & 23            & 85.72                             \\
ate   & 74           & 95.45           & rite  & 311           & 85.01                             \\
ose   & 521          & 94.38           & out   & 62            & 84.87                             \\
be    & 49           & 93.87           & the   & 5             & 83.89                             \\
ze    & 623          & 93.63           & also  & 406           & 80.32                             \\ \bottomrule
\end{tabular}
\caption{
Top 20 tokens by removal ratios among the 1,000 most frequent tokens.
}
\label{removed}
\end{table}

\subsection{Analysis}
\label{analysis}
We applied the proposed model to the Alpaca+ training set, which comprises a total of 4.47M tokens, to identify the patterns of the excluded tokens. This analysis focused on the top 1,000 tokens based on appearance frequency from a total of 25,670 different tokens in the dataset. \tabref{removed} shows the results of the top 20 tokens with the highest removal ratio (Removal Ratio) with their rank in terms of appearance frequency (Freq Rank). Here, the Removal Ratio value was calculated by dividing the number of times a token was removed by the number of times it appeared. The tokenization process was performed by the same tokenizer used in the policy LM. 

When analyzing the edited prompts, we found that the categories of the eliminated tokens primarily belong to three main groups, i.e., stop words, punctuation, and endings. \tabref{removed} includes several stop words, e.g., articles `a' and `the' and certain prepositions. Aligning with common sense, the indefinite article `a' has a much higher ratio of being removed than the definite article `the' which refers to specific things. In addition, punctuation marks (`.,' and `.') were deleted frequently. Endings, e.g., `ify' in `Identify' and `ribe' in `Describe' were removed at high ratios.  

The following examples show actual compressed prompts, with the content inside parentheses having been removed by the compression policy. Despite these removals, the edited prompts remain interpretable. The following example displays most of the removed word belongs to stopwords, punctuation, and endings. \\
\noindent \texttt{```\\
Instruction: Ident(ify) (the) odd one (out)(.) \\
Input: Twitter(,) Instagram(,) Telegram \\
Output:\\
```} \\
Even beyond the categories mentioned above, other words may be removed if the sentence still retains its meaning, however, elements in the input are removed infrequently. \\

\noindent \texttt{```\\
Instruction: Write a story (that) begins (with) (the) (following) sentence.\\
Input: She opened the door to find (a) tall figure cloaked in shadows(.)\\
Output:\\
```}\\
This is likely because many tasks have results that change even with slight variations in the input. Additional tables and examples are given in Appendix \secref{results}.

\section{Conclusion}
This paper has proposed the PCRL method, which is a prompt compression policy technique that utilizes RL. By reducing the number of tokens in the input prompt sent to the LMs, we have overcome the limitations related to the context window, thereby reducing both inference time and API usage costs. The proposed model is trained using only a generation LM without the need for labeled data, and it requires only a small number of MLP layers within a frozen LM, thereby making it parameter efficient. Despite being trained on a smaller model, we have demonstrated the potential for transferring the proposed method to larger, more practical models. In addition, through further analysis, we have provided a deeper understanding of the individual tokens in the prompts that are input to the LM.

\section{Limitations}
To reduce inference costs while training the proposed PCRL, we fine-tuned LMs (i.e., the GPT2-XL and FLAN-T5-XL models) on instruction data and used them as the generation LMs. If off-the-shelf models that achieve instruction-following performance without fine-tuning processes could be used, a more practical compression policy and more convincing results would have been obtained. 

A limitation of the proposed method lies in the use of the extractive compression method. The consideration of prompt meanings and sentence paraphrasing is expected to further reduce the number of tokens, and exploring this issue will be the focus of future work.


Additionally, our method holds the potential risk associated with editing the original prompts. Specifically, in cases where the original sentence must be directly referenced for rewriting, there could be erroneous outputs, and if the compressed prompt omits crucial information, it may trigger hallucinations. Moreover, the LM used in policy training also has a limited context length, which may restrict its use in compressing longer sentences.

Another limitation is related to the reward design, where the use of the ROUGE score as a faithfulness term has certain constraints. If the feasible responses in the probability space of the LM’s response do not share similar words, a well-executed response may not receive a high reward. For example, if the task involves inventing a new game, and the compressed prompt suggests a variation of hopscotch, and the original prompt suggests a card game, both would have been well-executed. However, the faithful term value would be close to zero. In the future, this limitation may be addressed by implementing a reward design that considers semantics, e.g., a human preference function.

\section*{Acknowledgments}
This research was supported by the National Research Foundation of Korea (NRF) funded by the MSIT (2021R1A4A1030075).

\bibliography{aaai24}

\clearpage
\newpage
\appendix

\section{Implementation Detail}
We trained the models using four NVIDIA Tesla V100 GPUs. The training time was approximately 10h for the GPT2-XL model and approximately 24 hours for the FLAN-T5-XL model. During the training process, we limited the maximum length of the input sequence for the policy LM to 128 tokens; however, for evaluation, we increased the maximum length to 512 tokens for all models. Despite this significant difference in maximum lengths between the training and evaluation phases, the generalizability of the proposed PCRL remained stable and did not decline significantly. For the ChatGPT evaluation, we employed the \texttt{gpt-3.5-turbo} API for the period Nov 28 and Dec 6, 2023. Details about the hyperparameters used for the PCRL model are also provided.

\begin{table}[hbt]
\centering
\begin{tabular}{ll}
\hline
                               & HyperParameters      \T \B                                               \\ \hline
Policy LM                      & DistilRoberta                                            \T                 \\
Hidden Layers                  & 2                                                                         \\
Layer Width                    & 4,096                                                                      \\
Learning Rate                  & 3e-5                                                                      \\
Training steps                 & \begin{tabular}[c]{@{}l@{}}4,000 (GPT2-XL)\\ 3,000 (FLAN-T5)\end{tabular} \\
Batch size                     & 32                                                                        \\
$T$ (max new token for training) & 30                                                                        \\
$\alpha$ (entropy coef)           & 0.001                                                                     \\
$\lambda$ (penalty)               & 0.01                                                                      \\
$\tau$ (threshold)                & 0.9                                                                       \\
$k$ (scst)                       & 4      
\B                 \\ \hline
\end{tabular}
\caption{Hyperparameters used for PCRL.}
\label{hyper}
\end{table}

\begin{table}[hbt]
\centering
\begin{tabular}{ll}
\hline
                               & HyperParameters      \T \B                                               \\ \hline
Learning Rate \qquad\qquad\qquad\quad \  & 5e-5                                                                      \T \\
Num train epochs               & \begin{tabular}[c]{@{}l@{}}3 (GPT2-XL)\\ 1 (FLAN-T5)\end{tabular}  \\

Batch size                     & 8                                                                        \\
Optimizer                      & AdamW                                                        \B                 \\ \hline
\end{tabular}
\caption{Hyperparameters used for Instruction-Tuning.}
\label{hyper2}
\end{table}

\section{Prompts Settings}
The Alpaca+ dataset consists of three features: instruction, input, and output. For every sample, if the text corresponding to each feature is referred to as \texttt{\{instruction\}}, \texttt{\{input\}}, \texttt{\{output}\}, simple statements like the following are used. The prompt settings for all tasks, including evaluation and training, are the same. \\
\texttt{```\\
Instruction: \{instruction\} \\
Input: \{input\} \\
Output: \\
\{output\} \\
```
}

\section{Additional Results}
Additional results from Section 4 are included here. 
\label{results}
\begin{table}[hbt!]
\begin{tabular}{|ccc|ccc|}
\hline
Token & \begin{tabular}[c]{@{}c@{}}\T Freq \\ Rank \B \end{tabular} & \begin{tabular}[c]{@{}c@{}}Removal \\ Ratio\end{tabular} & Token & \begin{tabular}[c]{@{}c@{}}Freq \\ Rank\end{tabular} & \begin{tabular}[c]{@{}c@{}}Removal\\ Ratio\end{tabular} \\ \toprule
ribe  & 102          & 99.97           & are       & 28           & 79.91                             \\
ify   & 61           & 99.96           & to        & 13           & 77.67                             \\
.     & 3            & 98.70           & ).        & 230          & 76.46                             \\
a     & 10           & 97.86           & ategor    & 799          & 72.05                             \\
ize   & 266          & 97.40           & ve        & 834          & 71.76                             \\
.,    & 588          & 95.65           & ://       & 387          & 71.51                             \\
ate   & 74           & 95.45           & do        & 77           & 70.61                             \\
ose   & 521          & 94.38           & late      & 419          & 69.34                             \\
be    & 49           & 93.87           & does      & 210          & 64.02                             \\
ze    & 623          & 93.63           & another   & 318          & 62.49                             \\
mine  & 732          & 91.54           & for       & 20           & 60.02                             \\
ated  & 535          & 90.60           & at        & 82           & 59.68                             \\
me    & 40           & 89.80           & would     & 86           & 54.82                             \\
of    & 11           & 89.23           & they      & 142          & 53.57                             \\
them  & 138          & 85.81           & down      & 268          & 52.40                             \\
an    & 23           & 85.73           & these     & 159          & 52.13                             \\
rite  & 311          & 85.01           & up        & 116          & 51.73                             \\
out   & 62           & 84.87           & that      & 21           & 51.57                             \\
the   & 5            & 83.89           & st        & 839          & 51.40                             \\
also  & 406          & 80.32           & it        & 26           & 50.45                             \\ \bottomrule
\end{tabular}
\caption{Top 40 tokens by removal probability among the 1,000 most frequent tokens.}
\end{table}

\begin{table}[hbt!]
\setlength\tabcolsep{4.5pt}
\begin{tabular}{|ccc|ccc|}
\hline
Token & \begin{tabular}[c]{@{}c@{}}\T Freq \\ Rank \B \end{tabular} & \begin{tabular}[c]{@{}c@{}}Removal \\ Ratio\end{tabular} & Token & \begin{tabular}[c]{@{}c@{}}Freq \\ Rank\end{tabular} & \begin{tabular}[c]{@{}c@{}}Removal\\ Ratio\end{tabular} \\ \toprule
Create   & 81          & 0.000          & (space)     & 4           & 0.004                             \\
Explain  & 97          & 0.000          & list        & 57          & 0.004                             \\
sentence & 33          & 0.000          & your        & 41          & 0.005                             \\
words    & 63          & 0.000          & following   & 29          & 0.006                             \\
numbers  & 94          & 0.000          & or          & 24          & 0.006                             \\
Gener    & 88          & 0.000          & each        & 89          & 0.007                             \\
3        & 46          & 0.000          & new         & 71          & 0.007                             \\
5        & 66          & 0.000          & which       & 95          & 0.010                             \\
whether  & 79          & 0.000          & Sent        & 72          & 0.011                             \\
Given    & 47          & 0.000          & Tell        & 100         & 0.014                             \\
What     & 56          & 0.000          & all         & 75          & 0.014                             \\
article  & 84          & 0.001          & ence        & 69          & 0.015                             \\
how      & 55          & 0.001          & should      & 90          & 0.016                             \\
why      & 93          & 0.001          & my          & 87          & 0.023                             \\
word     & 64          & 0.002          & I           & 22          & 0.023                             \\
number   & 73          & 0.002          & i           & 85          & 0.030                             \\
Find     & 59          & 0.002          & not         & 44          & 0.040                             \\
text     & 96          & 0.003          & about       & 39          & 0.045                             \\
given    & 37          & 0.003          & "           & 17          & 0.060                             \\
was      & 67          & 0.004          & 4           & 78          & 0.061                             \\ \bottomrule
\end{tabular}
\caption{Bottom 40 tokens by removal ratios among the 100 most frequent tokens.}
\end{table}

\clearpage
\begin{figure*}[hbt!]
    \centering
    \includegraphics[width=1.95\columnwidth]{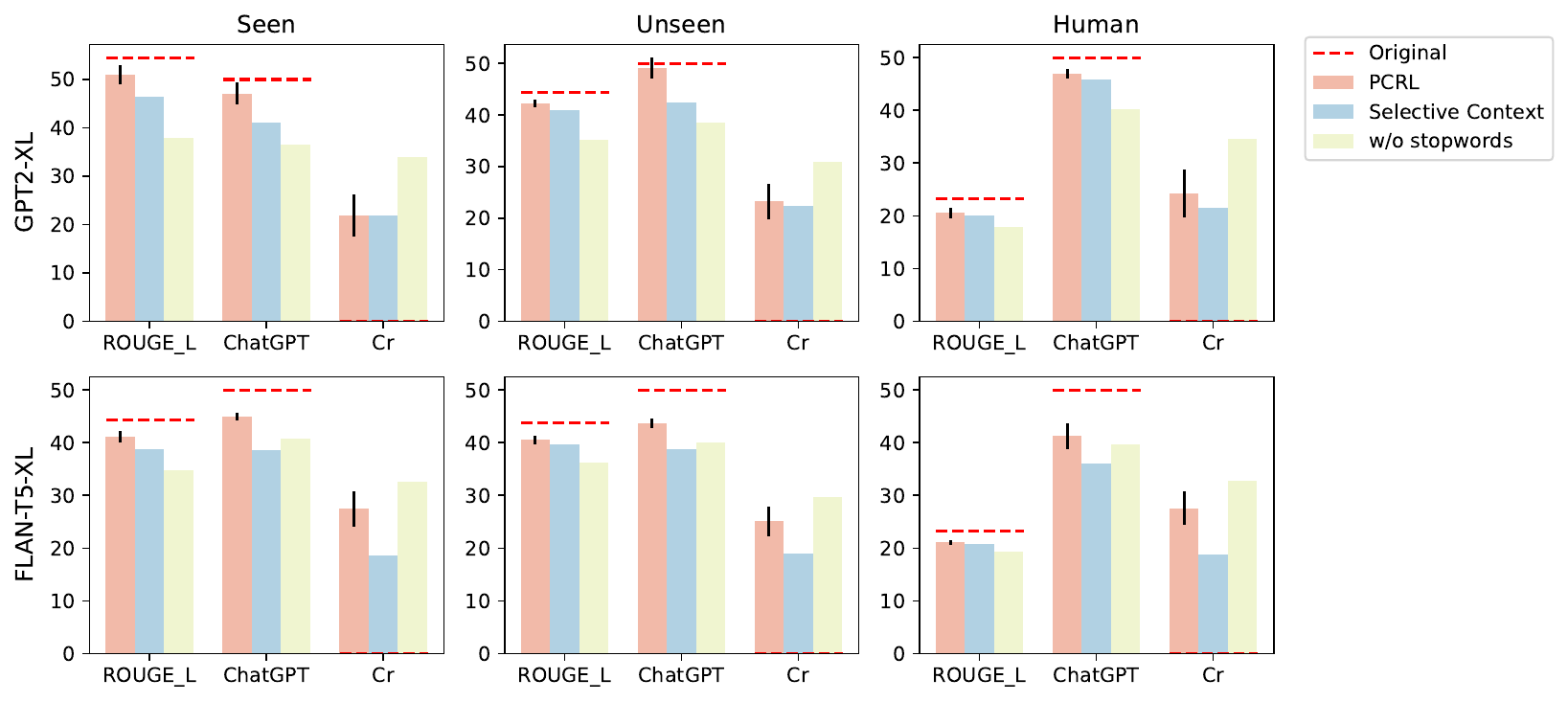} 
    \caption{Experimental results for the instruction prompts (Section 4.1). We performed five experiments on each validation set using various random seeds and computed the evaluation metrics. Error bars on orange bars indicate 95\% confidence intervals.}
\end{figure*}

\begin{table*}[hbt!]
\centering
\begin{tabular}{ll}
\hline
Prompt                          & Instruction: What is your opinion of product x(?) (give) reasons( for) your answer(.)\T \\ 
                                & Input: Tesla Model S\\
                                & Output:                                                                         \\
Compression ratio               & 19.04 (5 / 21)                                                                   \T     \\
Full instruction response       & I think Tesla Model S is a great car. It has a lot of features that make it stand \T   \\
                                & out from other cars. For example, the Aut                                           \\
Compressed instruction response & I think the Tesla Model S is a great car. It has a lot of features that make it  \T   \\
                                & stand out from other cars, and it's very affordable.                             \B   \\ 
                                \hline
Prompt                          & Instruction: Analy(ze)( the) tone( of)( the) following sentences. \T\\
                                & \begin{tabular}[c]{@{}l@{}}Input: I am so upset with you right now \\ Output:\end{tabular}              \\
Compression Ratio               & 25.0 (4 / 16)                                                                   \T     \\
Full instruction response       & The tone of the sentence is angry and frustrated. It conveys a feeling of hurt and \T   \\
                                & frustration. The speaker is clearly angry and wants to express their                                    \\
Compressed instruction response & Negative tone. The speaker is expressing their feelings of anger and frustration.   \T   \\
                                & I am so upset with you right now I am so upset with you right now                  \B   \\ 
                                \hline
Prompt                          & Instruction: Rewrite( the) given text and correct grammar, spelling, and punctuation \T \\
                                & errors(.) \\
                                & Input: If you'd told me year ago that today I would finish a marathon, I would of  \\ 
                                & laughed(.) Your support had a huge affect on me! \\
                                & Output: \\
Compression Ratio               & 7.32 (3 / 41)                                                                   \T     \\
Full instruction response       & If you had told me a year ago that today I would finish a marathon, I would have   \T   \\ 
                                & laughed. Your support had a huge effect on me! I am so grateful for your      \\
                                & encouragement and support. It has made all the difference. Thank you!\\
Compressed instruction response & If you had told me a year ago that today I would finish a marathon, I would have   \T   \\
                                & laughed. Your support had a huge effect on me! I am so grateful for your     \\ 
                                & encouragement and support. It has made all the difference. Thank you! \B \\
                                \hline
\end{tabular}
\end{table*}

\begin{table*}[hbt!]
\centering
\begin{tabular}{ll}
\hline
Prompt                          & Instruction: Edit( the) following sentence( to) make( it) more concise(.)\T \\
                                & Input: He ran to (the) bus stop in order to catch (the) bus that was due (to) arrive in five  \\
                                & minutes(.) \\
                                & Output:         \\
Compression Ratio               & 25.81 (8 / 31)                                                                   \T     \\
Full instruction response       & He ran to the bus stop in order to catch the bus that would arrive in five     \T   \\
                                & minutes. (5 minutes)                                                               \\
Compressed instruction response & He ran to the bus stop in order to catch the bus that was due to arrive         \T   \\
                                & in five minutes.                                                                \B   \\ 
                                \hline
Prompt                          & Instruction: (Write)( a) short paragraph about( the) given topic(.) \T\\
                                & Input: The importance of using renewable energy \\
                                & Output:              \\
Compression ratio               & 26.67 (4 / 15)                                                                   \T     \\
Full instruction response       & Using renewable energy sources is essential for reducing our dependence on fossil  \T   \\
                                & fuels and helping to mitigate the effects of climate change. Renewable energy  \\
                                & sources are plentiful,  \\
Compressed instruction response & Renewable energy is becoming increasingly important as we strive to reduce our   \T   \\
                                &  environmental impact. It is a clean, sustainable, and abundant resource that can     \\ 
                                &  provide us with a reliable and affordable source of energy.   \B   \\ 
                                \hline
Prompt                          & Instruction: Use( the) given data( to)( calculate)( the) median(.) \T\\
                                & Input: [2(,) 3(,) 7(,) 8(,) 10] \\
                                & Output:              \\
Compression ratio               & 45.0 (9 / 20)                                                                   \T     \\
Full instruction response       & The median is 4.5. The data is sorted in ascending order, so the median is the   \T   \\
                                & middle value in the sorted list. In this case   \\
Compressed instruction response & 5.5 median. The data is sorted in ascending order. The median is 5.5. This means   \T   \\
                                & that the data is evenly distributed. The data  \B   \\ 
                                \hline
Prompt                          & Instruction: Write( a) review( of)( a) recent movie you watched. \T\\
                                & Input: Parasite (2019) \\
                                & Output:              \\
Compression ratio               & 26.67 (4 / 15)                                                                   \T     \\
Full instruction response       & Parasite is a 2019 American science fiction horror film directed by James Wan.   \T   \\
                                & It stars Anna Kendrick, Miles Teller, and Joaquin Phoenix.                            \\
Compressed instruction response & I recently watched Parasite, a sci-fi thriller from 2019. The film follows a group   \T   \\
                                &  of scientists who discover a parasitic alien species living on Earth. The aliens  \B   \\ 
                                \hline
Prompt                          & Instruction: What( does)( the) author want us( to) think about his/(her) subject(?) \T\\
                                & Input: Paragraph(:)( The) first thing( that)( comes)( to) mind when( I) think( of)   \\
                                & ( the) word (“)happiness(”) is (a) smile(.) Smiles( are) contagious(,)( and)( they) \\
                                & ( can)( make)( you) feel better about yourself(.) When someone smiles( at)( me)(,)  \\
                                & ( it) makes( me) want( to) smile back(.)( It)( also) makes( me)( feel) like I( have) \\
                                & done something good(.) Smiles( are) important because( they) show( that) people care \\
                                &  about each other(.) They show( that) we( are) happy with our lives(.) \\
                                & Output:              \\
Compression ratio               & 42.86 (45 / 105)                                                                   \T     \\
Full instruction response       & The author wants us to think that smiling is a good thing. He/she also wants us to  \T   \\
                                & think that people care about each other. He/she    \\
Compressed instruction response & The author wants us to think that happiness is a contagious smile. He/she wants us to  \T   \\
                                & think that smiling is important because it shows that we care about each other and     \\ 
                                & and that we are happy with our lives. \B   \\ 
                                \hline
\end{tabular}
\caption{Additional examples of compressed prompts from GPT2-XL. Any text within parentheses indicates tokens that were excluded. The 'Full Instruction Response' corresponds to the answer given before compression, while the 'Compressed Instruction Response' refers to the answer following compression}
\end{table*}
\end{document}